\documentclass[letterpaper, 10 pt, conference]{IEEEtran}
\IEEEoverridecommandlockouts
\usepackage{cite}
\usepackage{amsmath,amssymb,amsfonts}
\usepackage{algorithmic}
\usepackage{graphicx}
\usepackage{textcomp}
\usepackage{cite}
\usepackage{multirow}
\usepackage[font=small]{caption}
\usepackage{subcaption}

\usepackage[table,xcdraw]{xcolor}
\usepackage{hyperref}

\def\BibTeX{{\rm B\kern-.05em{\sc i\kern-.025em b}\kern-.08em
    T\kern-.1667em\lower.7ex\hbox{E}\kern-.125emX}}
    
\makeatletter
    \newcommand{\linebreakand}{%
      \end{@IEEEauthorhalign}
      \hfill\mbox{}\par
      \mbox{}\hfill\begin{@IEEEauthorhalign}
    }
\makeatother

\title{Shake-VLA: Vision-Language-Action Model-Based System for Bimanual Robotic Manipulations and Liquid Mixing 
}

\author{
\IEEEauthorblockN{Muhamamd Haris Khan$^{*}$}
\IEEEauthorblockA{Skoltech\\
Moscow, Russia\\
Haris.khan@skoltech.ru}\\   
\and
\IEEEauthorblockN{Selamawit Asfaw$^{*}$}
\IEEEauthorblockA{Skoltech\\
Moscow, Russia\\
Selamawit.Asfaw@skoltech.ru}\\
\and
\IEEEauthorblockN{Dmitrii Iarchuk}
\IEEEauthorblockA{Skoltech\\
Moscow, Russia\\
Dmitrii.Iarchuk@skoltech.ru}\\     
\and
\IEEEauthorblockN{Miguel Altamirano Cabrera}
\IEEEauthorblockA{
Skoltech\\
Moscow, Russia\\
m.altamirano@skoltech.ru}\\
\linebreakand
\IEEEauthorblockN{Luis Moreno}
\IEEEauthorblockA{Skoltech\\
Moscow, Russia\\
Luis.Moreno@skoltech.ru}\\
\and
\IEEEauthorblockN{Issatay Tokmurziyev}
\IEEEauthorblockA{Skoltech\\
Moscow, Russia\\
Issatay.Tokmurziyev@skoltech.ru}
\and

\IEEEauthorblockN{Dzmitry Tsetserukou}
\IEEEauthorblockA{Skoltech\\
Moscow, Russia\\
d.tsetserukou@skoltech.ru}
\thanks{* These authors contributed equally to this work.}}

\begin{document}

\maketitle

\begin{abstract}

This paper introduces Shake-VLA, a Vision-Language-Action (VLA) model-based system designed to enable bimanual robotic manipulation for automated cocktail preparation. The system integrates a vision module for detecting ingredient bottles and reading labels, a speech-to-text module for interpreting user commands, and a language model to generate task-specific robotic instructions. Force Torque (FT) sensors are employed to precisely measure the quantity of liquid poured, ensuring accuracy in ingredient proportions during the mixing process. The system architecture includes a Retrieval-Augmented Generation (RAG) module for accessing and adapting recipes, an anomaly detection mechanism to address ingredient availability issues, and bimanual robotic arms for dexterous manipulation. Experimental evaluations demonstrated a high success rate across system components, with the speech-to-text module achieving a \textbf{93\%} success rate in noisy environments, the vision module attaining a \textbf{91\%} success rate in object and label detection in cluttered environment, the anomaly module successfully identified 95\% of discrepancies between detected ingredients and recipe requirements, and the system achieved an overall success rate of \textbf{100\%} in preparing cocktails, from recipe formulation to action generation.
\end{abstract}



\section{Introduction}

Human-robot interaction (HRI) has evolved significantly, particularly with the integration of advanced technologies that allow robots to understand and respond to human commands more intuitively. These sophisticated and interactive systems are capable of performing complex tasks in service-oriented domains. 

\begin{figure}[htp!]
  \centering
  \includegraphics[width=0.48\textwidth]{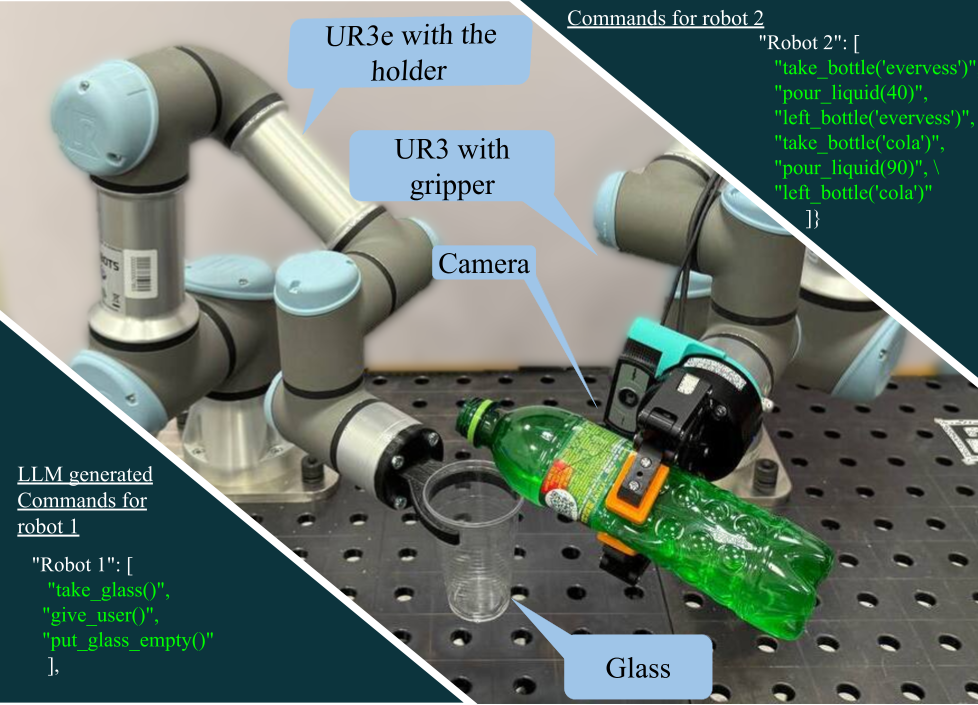}
  \caption{System Overview of Shake-VLA}
  \label{fig:device}
\end{figure}
The primary goal of HRI is to improve communication, coordination, and collaboration between humans and robots~\cite{sarker2024cohrtcollaborationhumanrobotteamwork},\cite{8581081}, \cite{ekrekli2023cospeechgestureshumanrobotcollaboration} leading to diverse applications ranging from industrial automation and healthcare to education and personal assistance~\cite{8678448},\cite{10.1145/3568162.3576993}, \cite{mishra2024personalized}. Ultimately, this seamless integration aims to create robots that can operate more effectively alongside humans~\cite{oleg2022cobottouch,rakhmatulin2021coboguider}.

As HRI progresses, the focus shifts towards enhancing robots' physical capabilities through coordinated dual-arm robotic systems~\cite{article}. These systems, which consist of two robotic arms working together, are designed to increase performance efficiency and handle tasks requiring precision, flexibility, and adaptability~\cite{inbook}. The need for such advanced manipulation techniques arises from the growing complexity of human environments and the demand for robots to execute intricate tasks autonomously.

To meet this demand, generative AI has emerged as a transformative force in robotics, enhancing autonomy, adaptability, and decision-making~\cite{openai2024gpt4technicalreport}. This technology plays a pivotal role in generating control policies and behavior models, enabling robots to learn and adapt to dynamic environments. The integration of vision, language, and action models further amplifies these capabilities, allowing robots to interact seamlessly with both humans and their surroundings.

For instance, CLIPort~\cite{shridhar2022cliport} demonstrates the potential of combining visual and language models to enhance robotic manipulation, translating human language into actionable robotic tasks. Similarly, RT-2~\cite{brohan2023rt2visionlanguageactionmodelstransfer} introduces vision-language integration, enabling robots to interpret visual inputs and commands, thereby improving generalization and complex reasoning. Bi-VLA~\cite{fidele2024bi} extends this by incorporating vision-language action models for bimanual robotic manipulation, using large language models (LLMs) to process user inputs and visual data to generate precise actions.

Moreover, the integration of continuous sensor data with language processing, as seen in PaLM-E~\cite{driess2023palm}, highlights the expanding scope of multimodal AI in robotic control. This convergence of HRI, robotic manipulation, and generative AI culminates in innovative solutions such as Industry 6.0~\cite{lykov2024industry60newgeneration}, where generative AI powers a swarm of heterogeneous robots capable of executing assembly tasks based on natural language descriptions provided by users. 
\begin{figure*}[t]
    \centering
    \includegraphics[width=\textwidth]{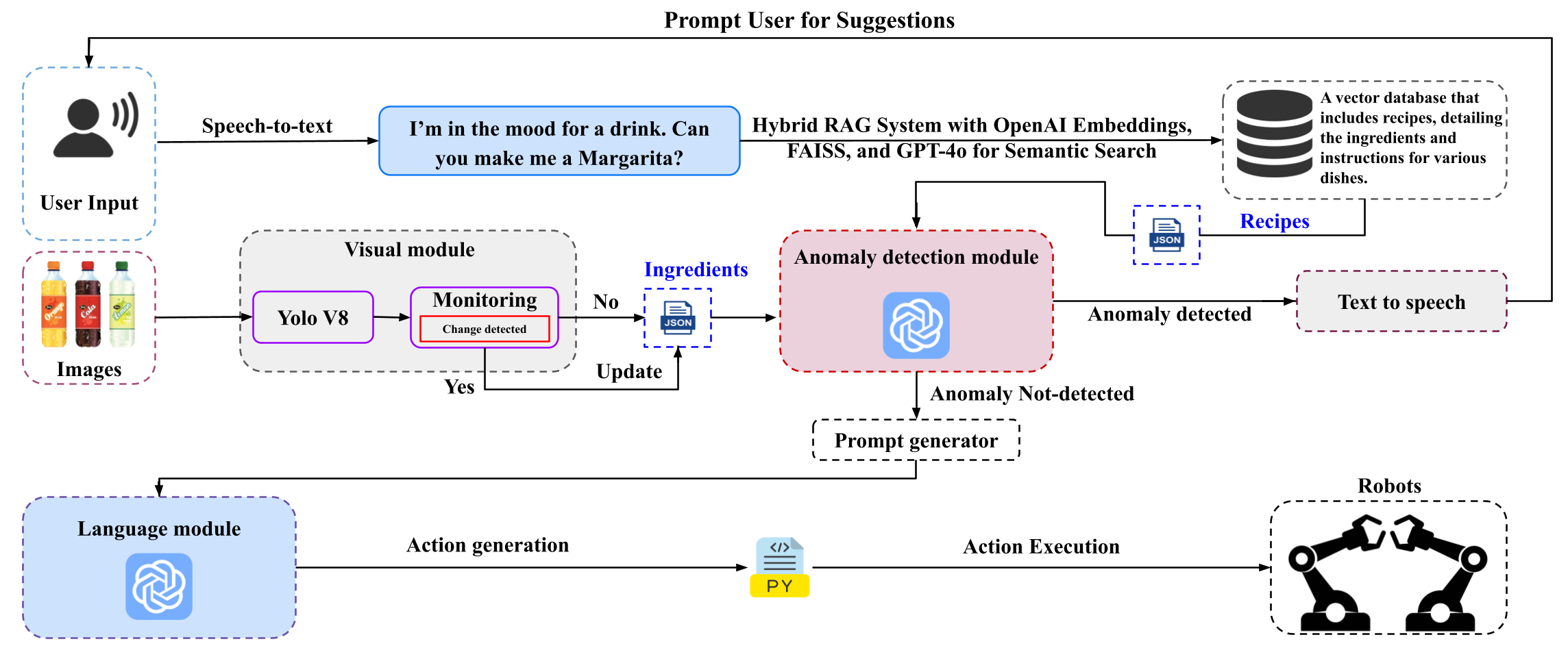} 
    \caption{Shake-VLA System Architecture.}
    \label{fig:system_architecture}
\end{figure*}
To build upon these advancements, this paper introduces \textbf{Shake-VLA}, a Vision-Language-Action (VLA) model-based system that leverages the seamless integration of vision, language, and action in robotic systems.
Shake-VLA shown in Fig.~\ref{fig:device} addresses the complex task of cocktail preparation through bimanual robotic manipulation, overcoming the dual challenges of precise bimanual control and naturalistic HRI. By advancing the state of the art in these areas, Shake-VLA showcases the potential of robots to perform complex, interactive tasks in real-world settings.

\section{System architecture} 
Fig.~\ref{fig:system_architecture} presents the system architecture, which simplifies automated decision-making in recipe preparation by integrating computer vision, speech processing, retrieval-augmented generation (RAG), anomaly detection, and language modeling—all in real-time. It understands user queries, detects ingredients through vision-based detection, and retrieves relevant recipes from a database. The system verifies ingredient availability, flags discrepancies, and offers user suggestions to resolve them. Generate step-by-step instructions for robots to execute the recipe, ensuring efficiency, adaptability, and user engagement.

\subsection{Visual module}
The visual module helps interpret the environment by locating objects and reading their text. It applies computer vision approaches on live video, generating structured data for later use. First, it captures video frames from a camera. Then, the YOLOV8 model spots objects and labels them with bounding boxes. When objects contain text, EasyOCR retrieves this content, revealing details such as names. All collected data, including labels, positions, and text, is stored in JSON format for further processing. The system also tracks any changes in the surroundings, updating the JSON to represent the current state. By bridging the real world with digital data, the module provides key inputs for tasks like matching objects to recipes and identifying issues. Combining object detection with text retrieval improves accuracy and adaptability across diverse contexts. This ensures continued effectiveness over time. New bottles can be dynamically integrated into the system as the visual module continuously updates the JSON representation when changes are detected, allowing seamless inclusion of new items into ongoing processes.

\subsection{speech-to-text and text-to-speech}

The system facilitates seamless interaction by converting spoken language into text and text into speech. Speech recognition is performed using the OpenAI Whisper-1 model \cite{radford2023robust}, which processes audio captured from a microphone and converts it into text for commands such as requesting recipes or troubleshooting.

For speech output, the system uses Google Text-to-Speech (gTTS) to convey information, transforming text into audio played through speakers. This enables the system to provide notifications, suggestions, or updates in real-time. Together, these components create a two-way voice communication interface, enhancing accessibility and usability by eliminating the need for written input or output while ensuring a smooth user experience.

\subsection{Retrieval Augmented Generation}

The Retrieval Augmented Generation (RAG) system retrieves information and forms answers for user queries. It has two parts: retrieval and generation. The retrieval stage uses FAISS (Facebook AI Similarity Search) for recipe embeddings built with OpenAI’s text-embedding-ada-002 model \cite{Greene2022Embedding}. When a user asks a question, it becomes a vector with the same embedding model. FAISS \cite{douze2024faiss} then finds the most relevant recipes. The chosen recipe goes to the generation stage, which uses GPT-4o to produce detailed answers, including step-by-step robotic instructions.

A hybrid RAG approach increases efficiency, scale, and accuracy by splitting retrieval and generation. FAISS enables quick searches on large datasets, while the language model focuses on creating responses. A single language model alone struggles with memory and context, making it less effective for handling changes in recipe data. In our case, the retrieval and generation process took a few milliseconds to a few seconds, depending on the complexity of the recipe.

The system maintains accuracy and adaptability through a vector database, allowing updates without retraining. Its modular design lets parts—such as the retrieval tool or language model—be replaced independently, ensuring precision in evolving contexts.

\subsection{Anomaly module}
The anomaly module ensures recipe requirements match available ingredients, keeping the mixing process on track even if problems arise. It takes two inputs: the recipe from the RAG system and the visual module’s list of ingredients, then compares them to find missing or mismatched items. For instance, if sugar and lime are required but only lime is detected, sugar is flagged as missing.

When that occurs, text-to-speech alerts the user to the missing ingredient and suggests alternatives like honey. The user’s choice is updated in the recipe. If no response is possible, predefined rules fix minor gaps automatically.

By managing ingredient shortages and considering user input, this module boosts reliability and flexibility. It supports smooth cooking, adapts to variations, and keeps recipes aligned with personal preferences, resolving inconsistencies for a trouble-free preparation process.

\subsection{Language module}
The language module is crucial for turning data and user inputs into clear instructions or answers. Using GPT-4o \cite{openai2024gpt4technicalreport}, it translates structured information—such as recipes and detected issues—into natural language for both users and robots.

When a user requests a recipe or addresses ingredient problems, the module generates step-by-step instructions, providing a precise sequence of robotic arm functions. This ensures that actions stay accurate and easy to perform.

The module receives prompts containing user questions, recipe details, and any anomaly module updates. These prompts give GPT-4o the necessary context to produce accurate, user-friendly outputs. For user interactions, it might respond, “Sugar is missing. Would you like to use honey?” That response is then relayed to text-to-speech for clear communication.

To enable action generation, an extensive set of API functions manages the robots’ operational capabilities. These functions handle a broad range of tasks, including:
\begin{itemize}
    \item \textbf{take\_glass()}: to pick up the empty glass.
    \item \textbf{take\_bottle(label)}: to pick up the targeted bottle.
    \item \textbf{left\_bottle(label)}: to left it on the table.
    \item \textbf{pour\_liquid(quantity, tolerance = 0.01)}: to pour liquid from bottle to the glass.
    \item \textbf{give\_user()}: to give glass to the user.
\end{itemize}

\begin{figure*}[t]
    \centering
    \includegraphics[width=\textwidth]{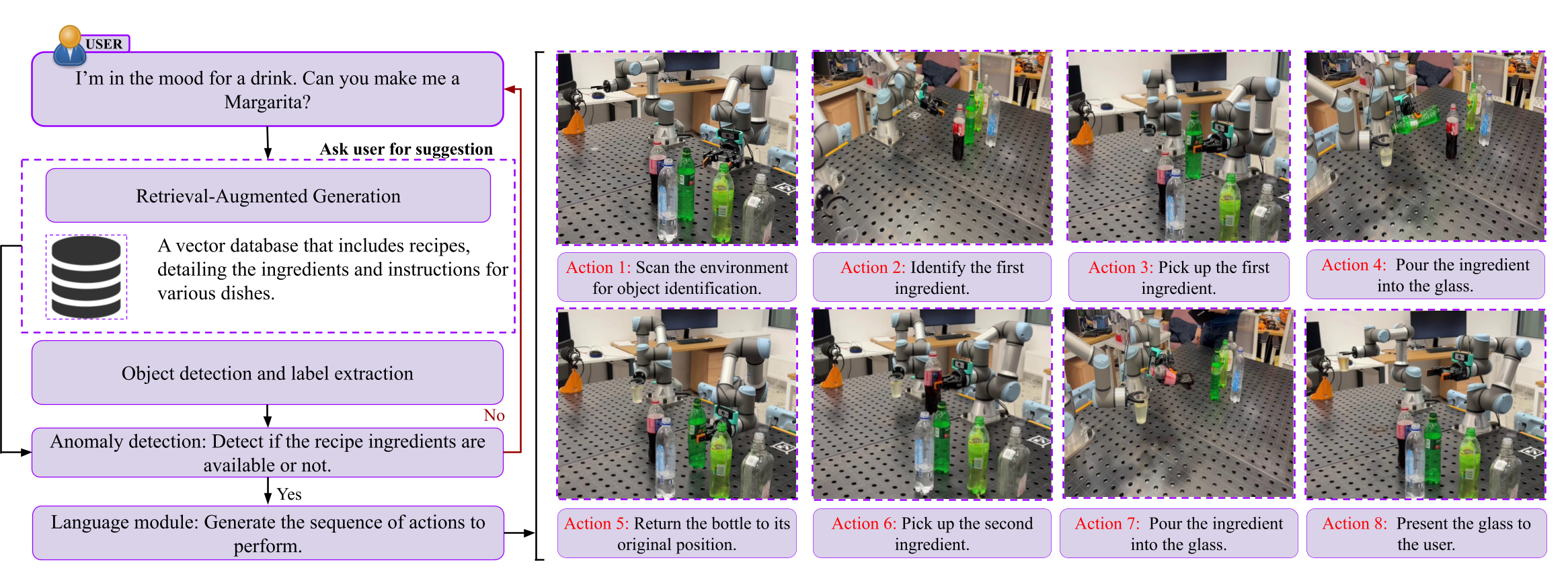} 
    \caption{Shake-VLA Cocktail Preparation Workflow.}
    \label{fig:workflow}
\end{figure*}

\section{Experimental evaluation} 

We ran a structured evaluation of Shake-VLA across four main parts: object detection, speech recognition, anomaly resolution, and system integration. Each was assessed for accuracy, reliability, and adaptability under realistic conditions, pinpointing strengths and areas that need more work.

The visual module, tasked with finding objects and reading their labels, was tested using 20 bottle setups with various sizes and labels in both Russian and English. Despite training solely in English, it achieved \textbf{91\%}, accuracy and correctly retrieved key details from Russian labels, forwarding clear information to later stages. Most errors arose with lengthy labels or those crowded by extra text. These results show the module’s effectiveness in cluttered scenarios. Yet, adding an OCR system tailored for multiple languages could further improve recognition and overall capability.

The speech-to-text module was evaluated for its ability to turn spoken commands into clear, actionable text—an essential link between user and system. Tests took place in noisy settings with diverse accents and varied speaking speeds. Each transcription was checked to ensure it remained understandable despite minor typos or grammar issues. A command was deemed successful if no extensive guesswork was needed by the language model. Of the \textbf{30} commands tested, \textbf{28} were accurately interpreted, yielding a \textbf{93\%} recognition rate. While these results confirm its reliability, advanced noise-cancellation methods could further boost performance under more challenging conditions.

Building upon the data processed by the visual and speech modules, the anomaly detection system was evaluated for its role in ensuring consistency between detected ingredients and recipe requirements. In a series of \textbf{20} trials conducted with varying orders, ingredients, and recipes, this module successfully identified \textbf{95\%} of discrepancies. These discrepancies included instances of missing or mismatched items. Additionally, the module dynamically proposed substitutions to ensure the continuity of the task. It effectively communicated with users to suggest alternatives, ensuring smooth task execution. However, the system faced difficulties in scenarios requiring ambiguous substitutions, pointing to the need for more sophisticated decision-making algorithms in such cases.

Finally, the fully integrated Shake-VLA system was tested in real-world scenarios to evaluate its overall reliability and adaptability. The setup included one UR3 robotic arm with a 2F-Robotiq gripper and one UR3e robotic arm, with a Logitech camera mounted on the UR3 for object detection and a cup holder attached to the UR3e for fluid handling. The visual module converted detected object bounding boxes from 2D to 3D coordinates, enabling precise determination of target item positions using methods described in \cite{fidele2024bi}. Additionally, a force sensor mounted on the UR3e measured the required fluid weight for recipe tasks. The system demonstrated an overall success rate of \textbf{100\%} in accurately preparing drinks, provided the recipe was retrieved successfully and the ingredients were available, showcasing its potential for practical applications in service-oriented domains. The complete workflow for cocktail preparation is illustrated 
Fig.~\ref{fig:workflow}.

These results collectively highlight the effectiveness of the Shake-VLA system while identifying opportunities for refinement, particularly in multilingual text processing, noise reduction, and anomaly resolution in edge cases.

\section{Conclusion and Future work}

This paper presents Shake-VLA, a system designed for bimanual robotic manipulation, specifically for automated cocktail preparation. The system combines various modules to address the challenges of precision manipulation and liquid mixing. Its architecture includes a vision module for detecting ingredient bottles and reading labels, a speech-to-text module for interpreting commands, a retrieval-augmented generation (RAG) module for recipe access and adaptation, and an anomaly detection module for resolving ingredient mismatches. A force-torque (FT) sensor ensures accurate liquid measurements during pouring. Experimental evaluations showed the speech-to-text module achieved a 93\% success rate in noisy environments, the vision module 91\% success in detecting objects and labels in cluttered conditions, and the anomaly module identified 95\% of recipe discrepancies. The integrated system achieved 100\% success in preparing cocktails, from recipe formulation to execution.

Despite these strong results, areas for improvement remain. Future work will focus on enhancing multilingual text recognition in the vision module to increase applicability in diverse settings. Improved noise-cancellation for the speech-to-text module could boost performance in noisy environments. Expanding the system's capabilities to include advanced robotic control for bimanual tasks beyond cocktail preparation—such as laboratory automation or pharmaceutical handling—is another promising direction. Incorporating adaptive learning to personalize recipes based on user feedback could further increase versatility. These refinements will strengthen Shake-VLA’s potential as a benchmark for service-oriented robotic systems.

\section*{Acknowledgements} 
Research reported in this publication was financially supported by the RSF grant No. 24-41-02039.
\bibliographystyle{IEEEtran}

\end{document}